\documentclass[letterpaper]{article} 
\usepackage{aaai23}  
\usepackage{times}  
\usepackage{helvet}  
\usepackage{courier}  
\usepackage[hyphens]{url}  
\usepackage{graphicx} 
\usepackage{natbib}  
\usepackage{caption} 
\usepackage{algorithm}
\usepackage{algorithmicx}
\usepackage{algpseudocode}

\usepackage{url}
\usepackage{graphicx}
\usepackage{subfigure}
\usepackage{amsmath,amsfonts}
\usepackage{bbm}

\usepackage{newfloat}
\usepackage{listings}
\DeclareCaptionStyle{ruled}{labelfont=normalfont,labelsep=colon,strut=off} 
\floatstyle{ruled}
\newfloat{listing}{tb}{lst}{}
\floatname{listing}{Listing}
\title{When Neural Networks Fail to Generalize? A Model Sensitivity Perspective}
\author{
    Jiajin Zhang\textsuperscript{\rm 1},
    Hanqing Chao\textsuperscript{\rm 1},
    Amit Dhurandhar\textsuperscript{\rm 2},
    Pin-Yu Chen\textsuperscript{\rm 2},
    Ali Tajer\textsuperscript{\rm 3},
    Yangyang Xu\textsuperscript{\rm 4},
    Pingkun Yan\textsuperscript{\rm 1} \footnote{Corresponding author} 
}
\affiliations{
    \textsuperscript{\rm 1}Department of Biomedical Engineering and Center for Biotechnology and Interdisciplinary Studies, Rensselaer Polytechnic Institute, Troy, NY, USA\\
    \textsuperscript{\rm 2}IBM Thomas J. Watson Research Center, Yorktown Heights, NY, USA\\
    \textsuperscript{\rm 3}Department of Electrical, Computer, and Systems Engineering, Rensselaer Polytechnic Institute, Troy, NY, USA\\
    \textsuperscript{\rm 4}Department of Mathematical Sciences, Rensselaer Polytechnic Institute, Troy, NY, USA\\

}

\usepackage{bibentry}

\begin{document}

\maketitle

\begin{abstract}
Domain generalization (DG) aims to train a model to perform well in unseen domains under different distributions. This paper considers a more realistic yet more challenging scenario, namely Single Domain Generalization (Single-DG), where only a single source domain is available for training. To tackle this challenge, we first try to understand \textit{when neural networks fail to generalize?} We empirically ascertain a property of a model that correlates strongly with its generalization that we coin as ``model sensitivity". Based on our analysis, we propose a novel strategy of Spectral Adversarial Data Augmentation (SADA) to generate augmented images targeted at the highly sensitive frequencies.
Models trained with these hard-to-learn samples can effectively suppress the sensitivity in the frequency space, which leads to improved generalization performance. 
Extensive experiments on multiple public datasets demonstrate the superiority of our approach, which surpasses the state-of-the-art single-DG methods by up to $2.55\%$. The source code is available at \url{https://github.com/DIAL-RPI/Spectral-Adversarial-Data-Augmentation}.
\end{abstract}

\section{Introduction}



Deep learning models may perform poorly when tested on samples drawn from out-of-distribution (OoD) data. 
Applications encountering OoD problems commonly involve natural domain shift~\cite{ben2010theory,pan2009survey} or image corruptions~\cite{hendrycks2016baseline, hendrycks2019benchmarking}. 
To tackle the problem of performance degradation in unseen domains, extensive research has been carried out on domain generalization (DG), which attempts to extend a model to unseen target domains by regularizing the model and exposing it to more data.


Based on how the source domain knowledge gets transferred to an unseen target domain, the existing DG techniques can be divided into three categories \cite{wang2022generalizing}, representation learning, constrained learning, and data manipulation.
The former two categories explicitly regularize a model to improve its generalizability. These approaches aim to learn domain invariant predictors by enhancing the correlations between the domain invariant representations and the labels. We would like to point out that the data augmentation based methods are actually also regularizing models, but implicitly. One of the contributions of our work is to visualize and quantify the effect of implicit regularization of data augmentation strategies.

Most of the existing DG methods learn the representations from multiple source domains~\cite{volpi2018generalizing, dou2019domain, muandet2013domain}. However, in many applications, there is only one single source domain available for training \cite{volpi2018generalizing, qiao2020learning, wang2021learning}. 
Despite the extensive literature on domain generalization, limited work deals with single source domain. 
In fact, many of the explicit regularization methods need multiple source domains to begin with and thus is inapplicable to this setting.
Data augmentation, as an effective strategy in deep learning, has shown promising performance in single domain generalization (single-DG) problems~\cite{volpi2018generalizing, xu2020robust, wang2021learning}. Such methods typically apply various operations to the source domain images to generate pseudo-novel domains~\cite{wang2021learning}. 
Models will be trained using both the source domain images and the augmented images with designed constraints to learn invariant representations.





Despite the popularity of data augmentation in single-DG, the existing methods bear two major drawbacks. They are either model agnostic or provide very limited OoD augmentation. 
Recent study~\cite{tan2021otce} on measuring cross-domain transferability demonstrated that the characteristics of both model and training data are important factors when quantifying the model's generalizability. However, the majority of data augmentation methods are model independent, which apply random generic image transformations to generate pseudo-domain images. Although they are helpful, those augmented images may not necessarily address the weaknesses of the models.
In contrast, recent methods exploiting adversarial samples for domain generalization~\cite{volpi2018generalizing, qiao2020learning,zhang2020robustified} learn to generate such augmentations by targeting the models' weakness. However, the resulted minor perturbations to the samples in the image space only trivially enhance the appearance diversity. Therefore, these adversarial samples based augmentation methods lead to limited generalization performance improvement. 


\begin{figure}[t]
    \centering
    \includegraphics[width=\columnwidth]{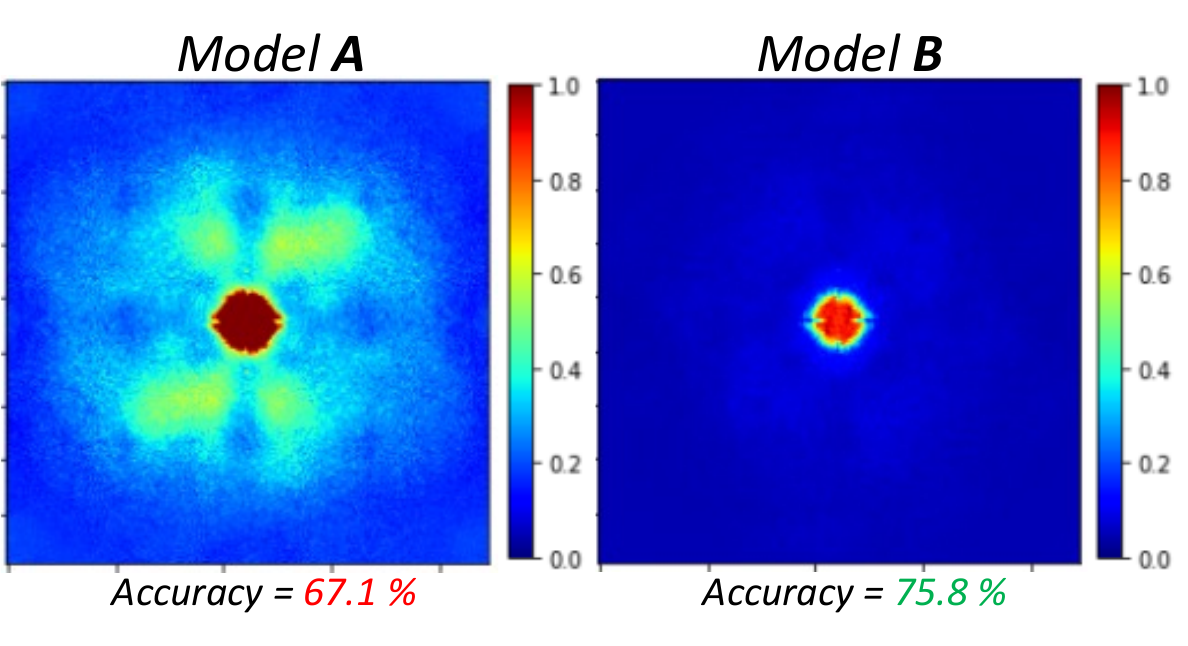}
    \caption{Introduction: we observed a clear correlation between the model generalization performance on an unseen target domain with the corresponding model sensitivity map.}
    \label{fig:intro_1}
\end{figure}

\begin{figure}[h]
    \centering
    \includegraphics[width=\columnwidth]{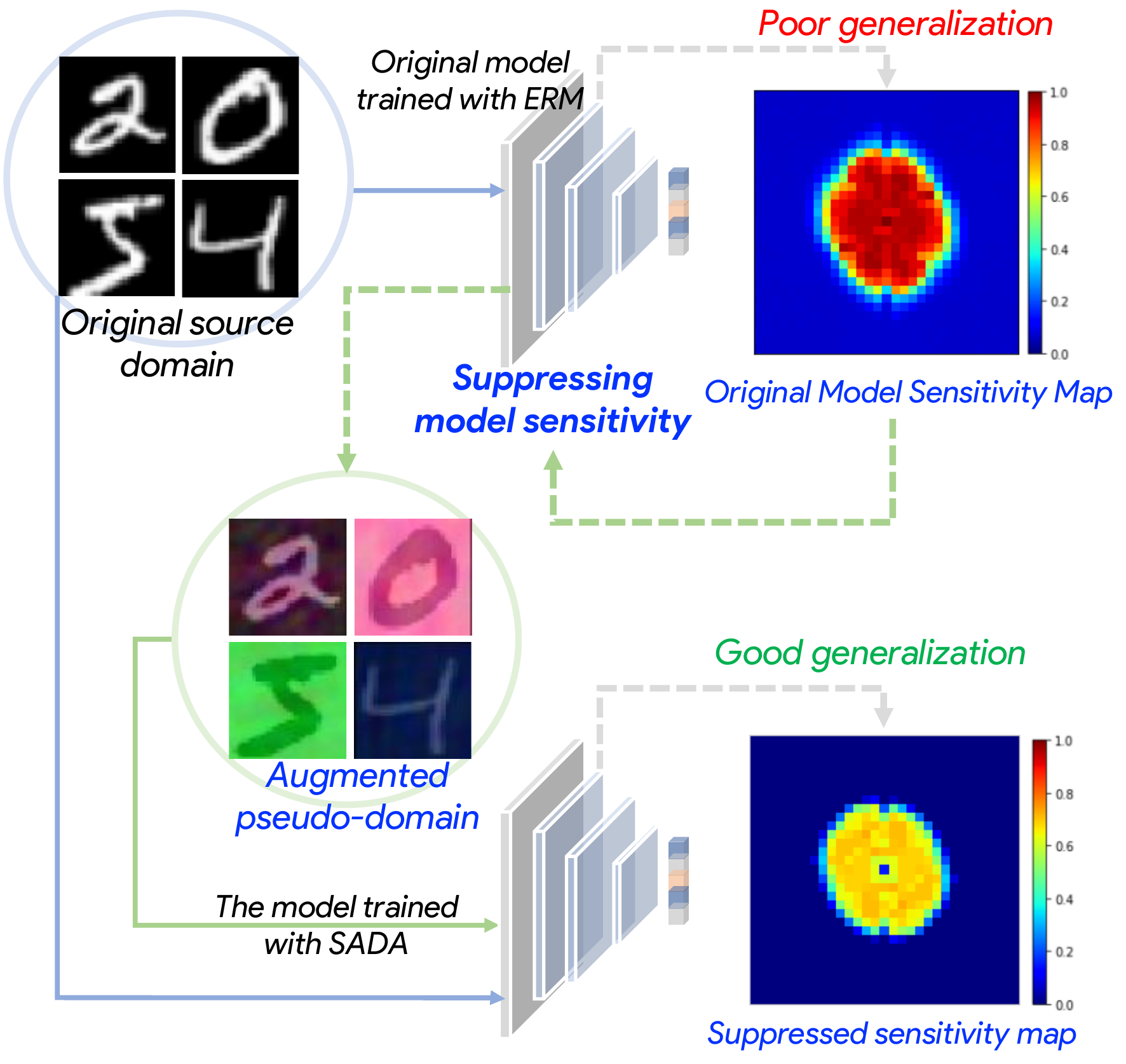}
    \caption{The overview of  our proposed Spectral adversarial data augmentation (SADA). The proposed model sensitivity map presents as a spectral indicator to quantify the model generalizability. Augmented pseudo-domain images are generated by SADA to boost the model performance by suppressing the source model sensitivity map.}
    \label{fig:intro_2}
\end{figure}

To tackle the above-mentioned challenges in data augmentation, we first ask a more fundamental question: \textit{when do neural networks fail in domain generalization?} In other words, we look into the model characteristics to quantify what aspects of the neural networks can reflect their generalizability. Inspired by the previous work on Fourier domain heatmap analysis~\cite{yin2019fourier}, we first propose a model sensitivity analysis approach to compute sensitivity maps as surrogates to help quantify the spectral weaknesses of the model. Fig.~\ref{fig:intro_1} presents two example sensitivity maps of two models sharing the same architecture but were trained using different strategies. The corresponding prediction accuracy of the two models on an unseen target domain shows that \textit{Model B} with less spectral sensitivity generalizes better than \textit{Model A}, which has much higher sensitivity. More detailed analysis and results on the sensitivity maps and their association with model generalizability are included in the experiment part of this article. 




The correlation between the generalization performance and model sensitivity map inspired us to design a novel data augmentation strategy to suppress the model sensitivity for improved single-DG.
We thus propose \textit{spectral adversarial data augmentation} (SADA), which curbs model sensitivity with targeted perturbation to the source domain data samples in the frequency space. Fig.~\ref{fig:intro_2} shows an overview of our framework. More specifically, we first train a model using the original source domain data through empirical risk minimization (ERM) and then compute the model sensitivity map. 
Since randomly augmenting images like in \cite{sun2021certified, xu2020robust, hendrycks2019augmix} may need generating a large number of images, the cost of data augmentation can be high and the following model training will be slow. To efficiently suppress the model sensitivity, instead of applying random operations, we target at each sensitive frequency point on the map and employ the adversarial techniques~\cite{zhang2022overlooked} to generate hard-to-learn samples. Such adversarial operation of the image amplitude spectrum allows us to largely augment samples with more appearance variation.
The generated samples are then mixed with the original samples to finetune the original model. 
%
%
Compared with other methods, SADA trained models present less sensitivity to domain shift across the frequency space, thus guarantee the better generalization performance. Based on such observation, we further develop a quantitative measure, which helps predict model generalizability.


The major contributions of this work are as follows.
\textit{1)} We introduce spectral sensitivity map as an indicator to quantify the model generalizability, which also visualizes the effect of implicit regularization such as data augmentation. \textit{2)} We propose SADA to improve the model generalization performance by suppressing the highly sensitive areas in the frequency space. SADA alleviates the drawbacks of the prior single-DG methods by targeting at model sensitivity and generating adversarial images with style variation. 
\textit{3)} We present thorough empirical analysis to compare the proposed method with the baselines from multiple perspectives on public datasets.

\section{Related work}
\label{related_work}

\subsection{Explicit regularization for DG}

One line of works on DG aims to train domain invariant classifier with explicit regularization~\cite{koyama2020invariance}.  A strategy that received significant attention in the last few years is invariant risk minimization (IRM) \cite{arjovsky2019invariant}. Given multiple environments, which correspond to different interventional distributions (viz. data from different sources) of a given data generating process, IRM promises to find invariant predictors that correspond to causal parents of a target variable. Efficient algorithms were designed \cite{ahuja2020invariant} and further analysis in support of the principle \cite{ahuja2021ermvsirm} have been done. However, it has been recently shown that the principle suffers from drawbacks in certain cases \cite{rosenfeldrisks2021,ahuja2021irminfbottleneck}, where it fails to uncover such predictors. 
Some studies adopt other strategies such as risk variance regularization~\cite{krueger2021out}, domains gradients alignments~\cite{koyama2020invariance}, smoothing cross domain interpolation paths~\cite{chuang2021fair}, and task-oriented  techniques~\cite{zhang2021task}. These approaches, however, generally require the target domain information, and cannot be directly adapted to the single-DG problem.

\subsection{Implicit regularization for DG} 
Data augmentation has been widely used to improve the generalization of deep learning models, which acts by implicitly regularization. Due to their effectiveness and simplicity, methods from the \textbf{*Mix*} family are the most commonly used approaches for data augmentation.
They augment data by mixing images with different random combinations, \textit{e.g.},
MixUp~\cite{zhang2017mixup}, CutMix~\cite{yun2019cutmix}, AugMix~\cite{hendrycks2019augmix}, PixMix~\cite{hendrycks2022pixmix}. In the single-DG settings, RandConv~\cite{xu2020robust} augments images with a random convolutional layer. L2D~\cite{wang2021learning} diversifies the image styles via mutual information maximization. However, most of the methods are model independent and thus the augmented images may not necessarily address the weaknesses of the models.

\textbf{Adversarial training} generates hard-to-learn samples targeted at the model weakness. To improve the single-DG performance, DUG~\cite{volpi2018generalizing} adversarially augments the representations of images to a fictitious domain. M-ADA~\cite{qiao2020learning} introduced a meta-learning framework to learn multi-adversarial domains with an autoencoder. AugMax~\cite{wang2021augmax} generates adversarial samples by selecting the worst-case weights of AugMix. However, the resulting minor perturbations in the image space only trivially enhance the appearance diversity. Thus, adversarial-based augmentation methods usually lead to limited generalization improvement. 

\textbf{Frequency spectrum augmentation} methods, including FDA~\cite{yang2020fda}, FDG~\cite{xu2021fourier} and FedDG~\cite{liu2021feddg}, generate images by either mixing up or swapping the low-frequency components of the source and target domain amplitude spectrum. Because of requiring target domain data, these methods cannot be directly adapted to the single-DG problem. To enhance the adversarial robustness under domain shift, FourierMix~\cite{sun2021certified} augments source images by adding noise to both amplitude and phase spectra. %
Our method instead, aiming to suppress the model spectral sensitivity, adversarially augments the image amplitude spectrum. We experimentally compare to typical baselines under a single-DG setting and demonstrate the superior performance.

\section{Methodology}
\label{methods}

The objective of single-DG is to train a model in one source domain, that can generalize well in many unseen target domains. We denote the source domain by $\boldsymbol{X}_S = \{(\boldsymbol{x}, \boldsymbol{y})\}$. $\boldsymbol{x} \in \mathbb{R}^{w \times h}$ is the source image, where $w$ and $h$ is the width and height. $\boldsymbol{y}$ is the corresponding label. As introduced in Fig~\ref{fig:intro_2}, to tackle this challenge, we propose the framework of \textit{spectral adversarial data augmentation} (SADA) to boost the model's generalizability by suppressing its spectral sensitivity. SADA first computes a model sensitivity map as a surrogate of the model vulnerability in the frequency space. Then it uses the model sensitivity map as guidance to synthesize spectral adversarial images, which encodes the model sensitivity into hard-to-learn augmentation images. In this section, the model sensitivity measurement and spectral adversarial augmentation processes are discussed in detail.

\subsection{Amplitude-modulated sensitivity map}
\label{model_sense}


To quantify the model's vulnerability/weakness to the different frequency corruptions, \cite{yin2019fourier} previously proposed the Fourier sensitivity analysis. Briefly, a \textit{Fourier basis} $\boldsymbol{A}_{i,j} \in \mathbb{R}^{w \times h}$ is defined as a Hermitian matrix with only two non-zero elements at $(i, j)$ and $(-i, -j)$, where the origin is at the image center.
%
A \textit{Fourier basis image} $\boldsymbol{U}_{i,j}$ is a real-valued matrix in the pixel space. It is defined as the $\ell_2$-normalized Inverse Fast Fourier Transform (IFFT) of $\boldsymbol{A}_{i,j}$, \textit{i.e.}, $\boldsymbol{U}_{i,j} = \frac{\mathcal{IFFT}(\boldsymbol{A}_{i,j})}{||\mathcal{IFFT}(\boldsymbol{A}_{i,j})||_2}$. Perturbed images are generated by adding the \textit{Fourier basis noise}
\begin{equation}
\label{eq:fourier_basis_noise}
\boldsymbol{N}_{i,j} = r\cdot\epsilon\cdot\boldsymbol{U}_{i,j} 
\end{equation}
to the original image $\boldsymbol{x}$ as $\boldsymbol{x} + \boldsymbol{N}_{i,j}$, where $\epsilon$ is a frequency-independent constant value to control the $\ell_2$-norm of the perturbation and $r$ is randomly sampled to be either -1 or 1. The \textit{Fourier basis noise} $\boldsymbol{N}_{i,j}$ only introduces perturbations at the frequency components $(i, j)$ and $(-i, -j)$ to the original images. The constant $\epsilon$ guarantees that images are uniformly perturbed across all frequency components. For RGB images, we add $\boldsymbol{N}_{i,j}$ to each channel independently following~\cite{yin2019fourier}.
The sensitivity at frequency $(i,j)$ of a given model $F$ trained on source domain is defined as the prediction error rate over the whole dataset $\boldsymbol{X}_S$:
%
\begin{equation}
\label{eq:sensitivity_map_org}
    \boldsymbol{M}_{org} (i,j) = 1 - \underset{\substack{(\boldsymbol{x},\boldsymbol{y}) \in \boldsymbol{X}_S}}{\rm {Acc}}(F( \boldsymbol{x} + \boldsymbol{N}_{i,j}, \boldsymbol{y})),
\end{equation}
where $\rm {Acc}$ is the model prediction accuracy.
By aggregating all the model sensitivity entries $\boldsymbol{M}_{org} (i,j)$ across the frequency space, a 2D model sensitivity map can be obtained as shown by the examples in Fig.~\ref{fig:intro_1}. The lowest frequency is at the center of the map and higher frequencies are closer to the edges.

Since $\epsilon$ is a frequency-independent constant, the original model sensitivity map defined by Eq.~\ref{eq:sensitivity_map_org} describes model's local vulnerability by uniformly perturbing all frequency components of the source images. Instead of a uniform distribution, the amplitude spectrum of natural images generally follows a power-law distribution~\cite{tolhurst1992amplitude}. Low-frequency amplitudes have much higher values than the high-frequency ones, and can vary more significantly across domains~\cite{yang2020fda}. Models generally would generalize poorly if such low-frequency variability is not presented in the training set~\cite{yang2020fda, liu2021feddg}. These observations indicate that the low-frequency components of images with large amplitude should be perturbed more significantly to truly reflect the model vulnerability \textit{w.r.t.} the domain shift problem. 


%
%



\begin{figure}[t]
    \centering
    \includegraphics[width=\columnwidth]{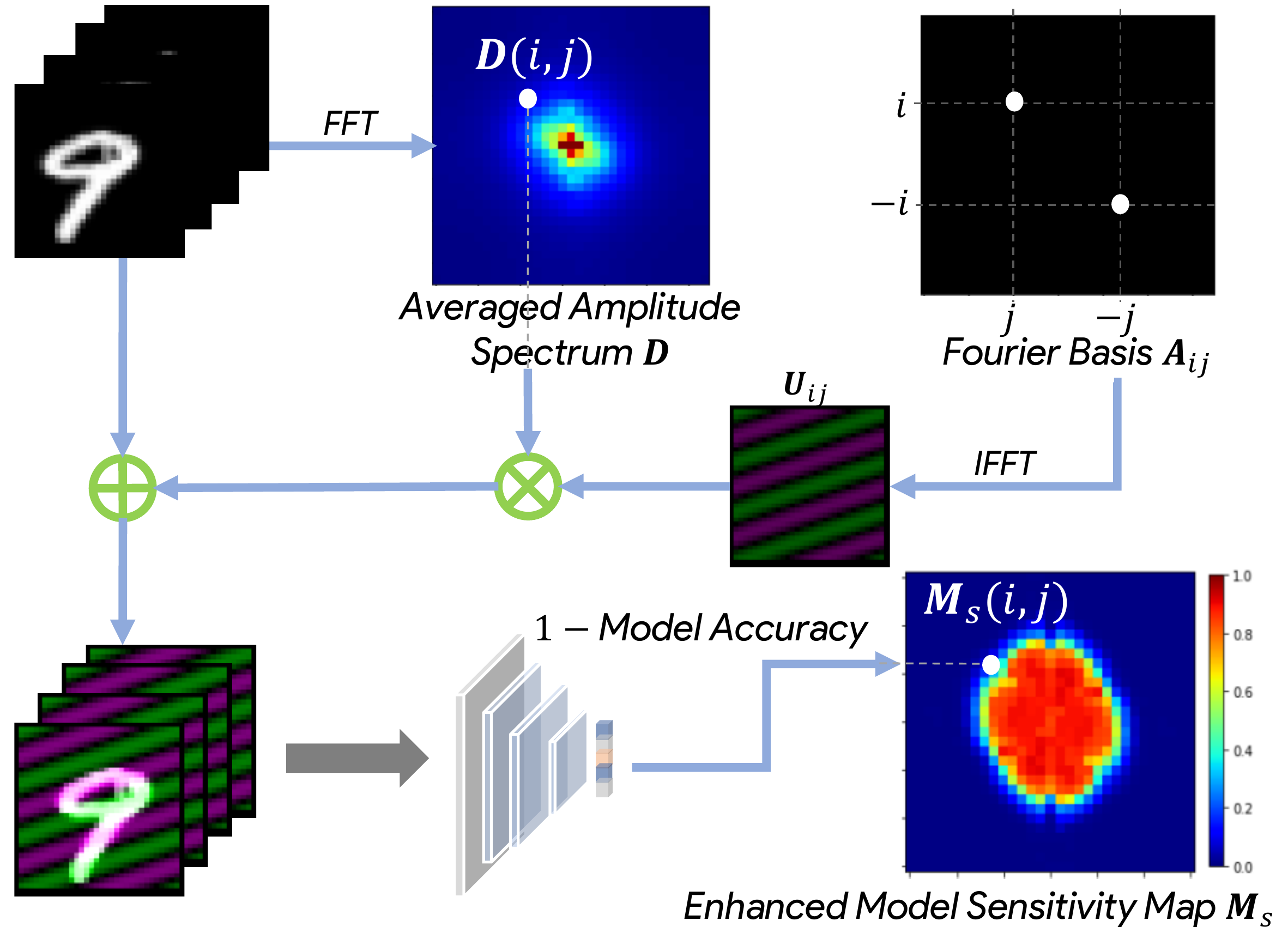}
    \caption{Enhancing model sensitivity map with source domain amplitude distribution. The averaged source amplitude spectrum $\boldsymbol{D}$ is encoded into the perturbed images.}
    \label{fig:method_smap}
\end{figure}

Thus, we propose to enhance the model sensitivity map by using the source domain amplitude spectrum as domain prior. As shown in Fig.~\ref{fig:method_smap}, a mean amplitude spectrum $\boldsymbol{D}$ is first computed by averaging the amplitude spectrum of all images in the source domain. Then, we reformulate the original \textit{Fourier basis noise} $\boldsymbol{N}_{i,j}$ by 
\begin{equation}
\label{eq:fourier_basis_noise_new}
\hat{\boldsymbol{N}}_{i,j} = r \cdot \boldsymbol{D} (i,j) \cdot \boldsymbol{U}_{i,j},
\end{equation}
where the frequency-independent $\epsilon$ in Eq.~\ref{eq:fourier_basis_noise} is replaced with the $(i, j)_{th}$ entry of $\boldsymbol{D}$ to control the noise level. 


Adopted from Eq.~\ref{eq:sensitivity_map_org}, the enhanced model sensitivity at frequency $(i, j)$ is computed by evaluating the prediction error rate on the perturbed source images as by
\begin{equation}
\label{eq:sensitivity_map_enhanced}
\boldsymbol{M}_S (i,j) = 1 - \underset{\substack{(\boldsymbol{x},\boldsymbol{y}) \in \boldsymbol{X}_S}}{\rm{Acc}}(F( \boldsymbol{x} + r \cdot \boldsymbol{D} (i,j) \cdot \boldsymbol{U}_{i,j} ), \boldsymbol{y}),
\end{equation}

\noindent where $F$ is a model trained with empirical risk minimization (ERM) by minimizing the cross entropy loss $\mathcal{L}_{ERM} =\underset{(\boldsymbol{x},\boldsymbol{y})\in\boldsymbol{X}_S}{\mathbbm{E}} \ell_{CE}(F(\boldsymbol{x}),\boldsymbol{y})$. In the experiment section, we quantitatively compared the enhanced model sensitivity map to the original one from different perspectives.



\subsection{Spectral adversarial data augmentation}



The model sensitivity $\boldsymbol{M}_S$ describes the model spectral weakness \textit{w.r.t.} the domain shift, which strongly correlates with the model cross-domain generalizability. 
The model sensitivity of certain frequencies can be suppressed if the diversity of the training data increases at those frequency elements.
Random spectral perturbation to the source images may help increase the overall diversity, however, generally lacks efficiency to sufficiently cover all potential pseudo domains.
Following this direction, we propose a \textit{spectral adversarial data augmentation} (SADA) method, which curbs model sensitivity with targeted perturbation to the source domain data samples in the spectral space. Instead of random transformation, SADA employs an adversarial technique to directly search for hard-to-learn samples by adding specially designed perturbations to the source images.
%


The entire pipeline of SADA is summarized in Alg.~\ref{alg:SADA}. More specifically, given a source domain image $\boldsymbol{x}$, its spectral amplitude $\boldsymbol{A}_{org}$ and phase $\boldsymbol{P}_{org}$ are computed by the Fast Fourier Transform (FFT) as 
\begin{equation}
    \label{eq:ifft}
    \boldsymbol{A}_{org}, \boldsymbol{P}_{org} = \mathcal{FFT}[\boldsymbol{x}].
\end{equation}
Then the original amplitude spectrum $\boldsymbol{A}_{org}$ is initialized with random perturbation as 
\begin{equation}
    \label{eq:randomize_amp}
    \boldsymbol{A}_{0} = \boldsymbol{A}_{org} \odot (1 + \rm Unif(-\epsilon, \epsilon)),
\end{equation}
where $\rm Unif(-\epsilon, \epsilon) \in \mathbb{R}^{w \times h}$ represents 2D matrix with each entry sampled uniformly from $[-\epsilon, \epsilon]$, and $\odot$ denotes the Hadamard product.

To target at each sensitive frequency component, as in Eq.~\ref{eq:update_amp}, the amplitude spectrum $\boldsymbol{A}_{t+1}$ is optimized iteratively by adding the $\boldsymbol{M}_S$-weighted sign gradient of the cross-entropy loss to the amplitude spectrum $\boldsymbol{A}_{t}$ with $\delta$ as the perturbation step size.

\begin{equation}
\label{eq:update_amp}
    \boldsymbol{A}_{t+1} = \boldsymbol{A}_{t} \cdot \{1 + \delta \cdot sign[\frac{\partial \ell_{CE}({F}(\mathcal{FFT}[\boldsymbol{A}_{t}, \boldsymbol{P}_{org}]), \boldsymbol{y})}{\partial \boldsymbol{A}_{t}}] \odot \boldsymbol{M}_S\}
\end{equation}

Previous studies~\cite{piotrowski1982demonstration, hansen2007structural, oppenheim1981importance, oppenheim1979phase} have demonstrated that the phase spectrum retains most of the semantic structure information of the original signals, while the amplitude mainly contains the style/domain-related statistics. Since the data augmentation objective is to diversify the image styles without affecting the original semantic meaning, we adversarially perturb the amplitude spectrum while keeping the original phase spectrum. That is, in each iteration, the augmented image is reconstructed from the updated amplitude $\boldsymbol{A}_{t+1}$ and the original phase spectrum $\boldsymbol{P}_{org}$. The reconstructed image is then clamped into the definition region $[0, 1]$ by $\boldsymbol{x}_{t+1} = \rm Clamp(\mathcal{FFT}[\boldsymbol{A}_{t+1}, \boldsymbol{P}_{org}], 0, 1)$.


\begin{algorithm}
    \caption{Spectral adversarial data augmentation.}
    \label{alg:SADA}
    \begin{algorithmic}[1]
        \Require{{\small{model $F$; source data $\{(\boldsymbol{x}_k, \boldsymbol{y}_k)\}_{k=1}^N$; initial perturbation level $\epsilon$; max steps $T$ and step size $\delta$; sensitivity map $\boldsymbol{M}_S$.}} 
        }
        \Ensure{augmented images $\{\Tilde{\boldsymbol{x}}_k\}_{k=1}^N$}
        
        \For{$k \leftarrow 1$ to $N$}
            \State {Compute the original spectrum $\boldsymbol{A}_{org}, \boldsymbol{P}_{org}$ by Eq.~\ref{eq:ifft}}
            
            \State {Randomly initialize amplitude $\boldsymbol{A}_{0}$ by Eq.~\ref{eq:randomize_amp}}
            
            \For{$t \leftarrow 0$ to $T$}
                \State{$\boldsymbol{x}_{k,t} \leftarrow \rm Clamp(\mathcal{IFFT}[\boldsymbol{A}_{t}, \boldsymbol{P}_{org}], 0, 1)$}
                \If {model prediction is changed by $\boldsymbol{x}_{k,t}$}
                    \State $break$ 
                    \Comment{Early stop for acceleration}
                \EndIf
                \State {Update amplitude spectrum $\boldsymbol{A}_{t}$ by Eq.~\ref{eq:update_amp}}
                \State{$\boldsymbol{A}_{t} = \rm Max(\boldsymbol{A}_{t} ,0)$} \Comment{constrain $\boldsymbol{A}_{t}>0$}
            \EndFor
            \State{$\Tilde{\boldsymbol{x}}_k \leftarrow \rm Clamp(\boldsymbol{x}_{k,t}, 0, 1)$}
        \EndFor
    \end{algorithmic}
\end{algorithm}



\subsection{Model training}
To learn invariant representations, we regularize the prediction consistency among the original image and all augmented images through a Jensen-Shannon ($\rm{JS}$) divergence~\cite{hendrycks2019augmix}. The total training loss is 
\begin{equation}
\label{eq:loss_tot}
    \mathcal{L} = \mathcal{L}_{ERM} + \lambda \cdot \rm{JS}(\boldsymbol{x}_{0}, \boldsymbol{x}_{1}, \boldsymbol{x}_{2}, ..., \boldsymbol{x}_{n}),
\end{equation}
where $\lambda$ is the trade-off parameter and $\boldsymbol{x}_{1}$, $\boldsymbol{x}_{2}$, $...\boldsymbol{x}_{n}$ are the $n$ augmented images from the same original image $\boldsymbol{x}_{0}$. The $\rm{JS}$ divergence is defined as $\rm{JS}(\boldsymbol{x}_{0}, \boldsymbol{x}_{1}, \boldsymbol{x}_{2}, ..., \boldsymbol{x}_{n}) = \frac{1}{n+1} \sum_{i=0}^n {\rm KL} ({F}(\boldsymbol{x}_{i}) || \Bar{\boldsymbol{p}})$, where $\rm KL$ is the Kullback-Leibler divergence, $F(\boldsymbol{x}_{i})$ is the model prediction probability of $\boldsymbol{x}_{i}$ and $\Bar{\boldsymbol{p}} = \frac{1}{n+1}\sum_{i=0}^n F(\boldsymbol{x}_{i})$.

\section{Experiments}
\label{experiments}

In this section, we conduct comprehensive experiments to evaluate SADA from different perspectives. Specifically, we aim to answer the following questions: \textbf{Q1}: Compared with prior methods, can SADA effectively improve the single-DG performance? (Sec.~\ref{sec:effectiveness}) \textbf{Q2}: Can the proposed model sensitivity map suggest when neural networks may generalize well? (Sec.~\ref{sec:sensitivity_perspective}) \textbf{Q3}: How is the data efficiency of SADA compared with other data augmentation methods? (Sec.~\ref{sec:efficiency})

\subsubsection{Datasets} 

To answer those questions, we evaluated our method and other benchmarks on three benchmark datasets.

1) \textbf{DIGITS} consists of 5 domains, including \textbf{MNIST}~\cite{lecun1998gradient}, \textbf{SVHN}~\cite{netzer2011reading}, \textbf{MNIST-M}~\cite{ganin2015unsupervised}, \textbf{SYNTH}~\cite{ganin2015unsupervised} and \textbf{USPS}~\cite{lecun1989backpropagation}. We converted all the gray scale images to RGB images. 

2) \textbf{PACS}~\cite{li2017deeper} is a more challenging domain generalization dataset including four domains, \textbf{P}hoto, \textbf{A}rt painting, \textbf{C}artoon, and \textbf{S}ketch. We follow the official dataset split for training validation and testing. 

3) \textbf{CIFAR-10-C}~\cite{hendrycks2019benchmarking} is the corrupted version of CIFAR-10~\cite{krizhevsky2009learning} by four categories of corruption, \textit{i.e.}, weather, blur, noise, and digital. Each corruption has $5$ level severity. 

\subsubsection{Implementation details} In all the experiments, we set the weighting factor $\lambda = 0.25$, perturbation steps $T = 5$, step size $\delta = 0.08$ and random initialization range $\epsilon = 0.2$. The number of augmented images per training sample in Eq.~\ref{eq:loss_tot} is set to $3$, based on our empirical evaluation results. 
We also include three SADA variants, 1SADA+2Mix, 2SADA+1Mix and 3SADA+0Mix, for comparison. The `\#' indicates the number of SADA and AugMix images included in the 3 augmented images per training sample.

For a fair comparison with other methods, we directly adopted the same network architectures of the previous works~\cite{volpi2018generalizing, qiao2020learning, wang2021learning}. For \textbf{DIGITS} dataset, we trained a ConvNet~\cite{lecun1998gradient} with SGD optimizer (default settings) for $50$ epochs. The initial learning rate is $0.001$, which decays by $0.1$ for every $20$ epochs. The batch size is $128$. 
For the \textbf{PACS} dataset, ResNet-18~\cite{he2016deep} is pretrained on Imagenet and finetuned in the source domain by SGD for $80$ epochs. The initial learning of $0.01$ is scheduled to decay by $0.1$ for every $20$ epochs. The batch size is $256$. 
For \textbf{CIFAR-10-C}, a Wide Residual Network~\cite{zagoruyko2016wide} with 16 layers and width of 4 (WRN-16-4) was optimized with SGD for $200$ epochs with batch size 256. The initial learning rate of $0.1$ linearly decays by $0.1$ for every $40$ epochs.

\subsection{Method effectiveness}
\label{sec:effectiveness}
We compared SADA with ERM, CCSA~\cite{motiian2017unified}, JiGen~\cite{carlucci2019domain}, d-SNE~\cite{xu2019d}, AugMix~\cite{hendrycks2019augmix}, GUD~\cite{volpi2018generalizing}, M-ADA~\cite{qiao2020learning}, RandConv~\cite{xu2020robust} and L2D~\cite{wang2021learning}. The same model architecture was used for all the approaches. 

\subsubsection{DIGITS} Table~\ref{tab:DIGITS_table} shows the 3-run average accuracy of all the methods trained on \textbf{MNIST} and evaluated in each target domain. All the variants of SADA achieved better accuracy than the baselines. Specifically, significant improvements of $5.50\%$ and $9.08\%$ are observed on the two very challenging target domains, \textbf{SVHN} and \textbf{SYNTH}, respectively. This performance gain mainly contributes to the spectrally augmented samples with large appearance/style variation. As we pointed out earlier, adversarial-based methods, such as GUD and M-ADA, generates only minor perturbations in the image space to enhance the appearance diversity, and thus couldn't outperform the random data augmentation methods, such as RandConv. 

\begin{table}[th]
	\centering
	\caption{3-run average accuracy and (standard deviation) of \textbf{MNIST}-trained models evaluated on \textbf{USPS}, \textbf{MNIST-M}, \textbf{SVHN}, and \textbf{SYNTH}.
	Best performance is in \textbf{bold}.}
	\scalebox{0.71}{
		\begin{tabular}{|l||c|c|c|c||c|}
			\hline
			& \multicolumn{4}{c||}{Target Domain} & \\
			\cline{2-5}
			\textbf{Method} &  \textbf{USPS} & \textbf{MNIST-M} & \textbf{SVHN} & \textbf{SYNTH} & Average\\
			\hline
			\hline
			ERM & 76.90\tiny{(0.34)} & 52.74\tiny{(0.23)}&  27.85\tiny{(0.16)} & 39.65\tiny{(0.22)} & 49.29\tiny{(0.22)}\\
			CCSA & 83.72\tiny{(0.68)} & 49.29\tiny{(0.82)}& 25.89\tiny{(1.12)}& 37.31\tiny{(0.86)}&  49.05\tiny{(0.78)}\\
            JiGen & 77.16\tiny{(1.12)} & 57.80\tiny{(0.72)} & 33.81\tiny{(0.75)} & 43.79\tiny{(1.67)} & 53.14\tiny{(0.85)}\\
            d-SNE & \textbf{93.16}\tiny{(0.63)} &  50.98\tiny{(0.97)} & 26.22\tiny{(0.89)} & 37.83\tiny{(0.77)} & 52.05\tiny{(0.72)}\\
            AugMix & 80.24\tiny{(1.27)} & 75.86\tiny{(0.84)}& 63.85\tiny{(0.79)}& 69.84\tiny{(1.04)}& 72.45\tiny{(0.77)}\\
			GUD & 77.26\tiny{(0.77)} & 60.41\tiny{(0.63)} & 35.51\tiny{(0.54)}& 45.32\tiny{(0.67)}& 55.67\tiny{(0.61)}\\
			M-ADA & 78.53\tiny{(0.56)} & 67.94\tiny{(0.73)} & 42.55\tiny{(0.81)} & 48.95\tiny{(0.97)} & 59.49\tiny{(0.69)}\\
			RandConv & 84.37\tiny{(0.90)} & \textbf{87.77}\tiny{(0.85)}& 57.56\tiny{(1.67)} & 62.85\tiny{(0.77)}& 72.88\tiny{(0.56)} \\
			L2D & 83.95\tiny{(0.77)} & 87.32\tiny{(0.91)} & 62.85\tiny{(0.78)} & 63.72\tiny{(0.74)} & 74.45\tiny{(0.70)}\\
			\hline
			\hline
			1SADA+2Mix & 81.92\tiny{(1.05)} & 80.88\tiny{(0.78)} & 67.66\tiny{(0.56)} & 70.65\tiny{(0.82)} & 75.28\tiny{(0.68)}\\
			2SADA+1Mix & 89.34\tiny{(0.92)} & 75.74\tiny{(0.70)} & 68.34\tiny{(0.61)} & 72.10\tiny{(0.77)} & 76.38\tiny{(0.69)}\\
			3SADA+0Mix & 89.29\tiny{(0.80)} & 75.61\tiny{(0.66)} & \textbf{68.45}\tiny{(0.67)}& \textbf{72.90}\tiny{(0.72)} & \textbf{76.56}\tiny{(0.65)}\\
			\hline
		\end{tabular}
	}
	\label{tab:DIGITS_table}
\end{table}

\subsubsection{PACS} 
We train a model in a single source domain and test on the other three target domains. The averaged accuracy on the three target domains are reported in Table~\ref{tab:PACS_table}. 
The proposed SADA variants achieve the best performance in 3 out of the 4 source domains, i.e., \textbf{P}hoto, \textbf{A}art, and \textbf{S}ketch. Both 2SADA+1Mix and 3SADA+0Mix achieved over $3.9\%$ improvement with \textbf{S}ketch as the source domain, which contains the largest domain shift from the other three colored domains. In addition, 2SADA+1Mix and 3SADA+0Mix consistently outperform 1SADA+2Mix, which indicates the importance of SADA augmented images. These observations verified that model generalization performance can improve, if SADA is included for suppressing the model's sensitivity.

\begin{table}[t!]
	\centering
	\caption{3-run average accuracy and (standard deviation) of models trained in each single domain (\textbf{P}hoto, \textbf{A}rt, \textbf{C}atoon, \textbf{S}ketch). Best performance is in \textbf{bold}.}
	\scalebox{0.73}{
		\begin{tabular}{|l||c|c|c|c||c|}
			\hline
			& \multicolumn{4}{c||}{Source Domain} & \\
			\cline{2-5}
			\textbf{Method} &  \textbf{P}hoto & \textbf{A}rt & \textbf{C}atoon & \textbf{S}ketch & Average\\
			\hline
			\hline
			ERM & 33.52\tiny{(0.47)} & 57.86\tiny{(0.43)}&  67.84\tiny{(0.51)} & 25.12\tiny{(0.40)} & 46.09\tiny{(0.32)}\\
			CCSA & 42.77\tiny{(0.81)} & 61.89\tiny{(1.02)}& 67.46\tiny{(0.97)}& 26.43\tiny{(0.83)}&  51.08\tiny{(0.75)}\\
            JiGen & 43.49\tiny{(0.94)} & 63.66\tiny{(0.84)} & 70.08\tiny{(0.77)} & 32.47\tiny{(1.12)} & 52.43\tiny{(0.79)}\\
            d-SNE & 46.28\tiny{(0.66)} &  63.20\tiny{(0.97)} & 26.22\tiny{(0.89)} & 37.83\tiny{(0.77)} & 52.05\tiny{(0.72)}\\
            AugMix & 48.27\tiny{(1.12)} & 72.92\tiny{(0.81)} & 73.81\tiny{(0.71)}& 54.88\tiny{(1.21)}& 62.47\tiny{(0.82)}\\
			GUD & 45.62\tiny{(0.81)} & 69.47\tiny{(0.83)} & 73.46\tiny{(0.68)}& 41.67\tiny{(0.82)}& 57.56\tiny{(0.72)}\\
			M-ADA & 48.22\tiny{(0.68)} & 70.46\tiny{(0.77)} & 75.67\tiny{(0.78)} & 43.26\tiny{(1.11)} &  59.40\tiny{(0.69)}\\
			RandConv & 50.86\tiny{(0.86)} & 75.82\tiny{(0.80)}& 75.46\tiny{(1.05)} & 48.90\tiny{(0.85)}& 62.76\tiny{(0.69)} \\
			L2D & 51.17\tiny{(0.77)} & 76.90\tiny{(0.97)} & \textbf{77.80}\tiny{(0.79)} & 53.68\tiny{(0.94)} & 64.74\tiny{(0.88)}\\
			\hline
			\hline
			1SADA+2Mix & 51.26\tiny{(0.79)} & 76.98\tiny{(0.78)}& 76.26\tiny{(0.62)}& 55.91\tiny{(0.78)}& 65.20\tiny{(0.62)}\\
			2SADA+1Mix & \textbf{51.22}\tiny{(0.84)} & \textbf{77.82}\tiny{(0.79)}& 76.94 \tiny{(0.91)}& \textbf{57.76}\tiny{(0.73)}& \textbf{66.18}\tiny{(0.74)}\\
			3SADA+0Mix & 51.18\tiny{(0.81)} & 77.68\tiny{(0.76)} & 76.35\tiny{(0.87)}& 57.61\tiny{(0.72)} & 65.71\tiny{(0.68)}\\
			\hline
		\end{tabular}
	}
	\label{tab:PACS_table}
\end{table}

\subsubsection{CIFAR-10-C}
Besides the natural domain shift in \textbf{DIGITS} and \textbf{PACS}, we further evaluated the method on the image corruption dataset. Table~\ref{tab:CIFAR10_table} shows the average accuracy of all the methods trained on \textbf{CIFAR-10} and evaluated on four types of corruption under the severest level 5. The averages accuracy of 2SADA+1Mix surpasses the best baseline AugMix by $2.55\%$. More detailed performance of all five-level corruptions are included in appendix, where the proposed SADA consistently outperforms other baseline methods at different severity levels. The results validate that SADA not only handles the natural domain shift but is resilient to artificial corruptions.

\begin{table}[th]
	\centering
	\caption{3-run average accuracy and (standard deviation) of models trained on \textbf{CIFAR-10} and evaluated on \textbf{CIFAR-10-C} dataset. Best performance is in \textbf{bold}.}
	\scalebox{0.73}{
		\begin{tabular}{|l||c|c|c|c||c|}
			\hline
			& \multicolumn{4}{c||}{Corruption Category} & \\
			\cline{2-5}
			\textbf{Method} &  \textbf{Weather} & \textbf{Blur} & \textbf{Noise} & \textbf{Digits} & Average\\
			\hline
			\hline
			ERM & 67.21\tiny{(0.66)} & 56.73\tiny{(0.41)}&  30.26\tiny{(0.42)} & 62.30\tiny{(0.35)} & 54.08\tiny{(0.23)}\\
			CCSA & 67.66\tiny{(0.74)} & 57.81\tiny{(0.86)}& 28.73\tiny{(0.97)}& 61.96 \tiny{(0.88)}&  54.04\tiny{(0.70)}\\
            JiGen & 67.20\tiny{(0.92)} & 58.06\tiny{(0.79)} & 30.37\tiny{(0.69)} & 62.05 \tiny{(1.10)} & 54.43\tiny{(0.69)}\\
            d-SNE & 67.90\tiny{(0.56)} &  56.59\tiny{(1.02)} & 33.97\tiny{(0.55)} & 61.83 \tiny{(0.75)} & 55.07\tiny{(0.61)}\\
            AugMix & 78.53\tiny{(1.12)} & 82.04\tiny{(0.81)}& 64.45\tiny{(0.71)} & 76.17\tiny{(1.21)}& 75.28\tiny{(0.82)}\\
			GUD & 69.94\tiny{(0.88)} & 60.57\tiny{(0.73)} & 48.66\tiny{(0.85)}& 60.37\tiny{(0.94)}& 59.91\tiny{(0.87)}\\
			M-ADA & 75.54\tiny{(0.68)} & 63.76\tiny{(0.77)} & 54.21\tiny{(0.78)} & 65.10 \tiny{(1.11)} & 64.65\tiny{(0.69)}\\
			RandConv & 76.87\tiny{(0.86)} & 55.36\tiny{(0.80)}& \textbf{75.19}\tiny{(1.05)} & 77.51\tiny{(0.85)}& 71.23\tiny{(0.69)} \\
			L2D & 75.98\tiny{(0.77)} & 70.21\tiny{(0.97)} & 73.29\tiny{(0.79)} & 72.02\tiny{(0.94)} & 72.88\tiny{(0.88)}\\
			\hline
			\hline
			1SADA+2Mix & 78.69\tiny{(0.89)} & 82.10\tiny{(0.81)} & 67.95\tiny{(0.87)}& 77.32\tiny{(0.91)}& 75.52\tiny{(0.82)}\\
			2SADA+1Mix & 79.14\tiny{(0.81)} & \textbf{82.38}\tiny{(0.76)} & 71.42\tiny{(0.87)}& 78.38\tiny{(0.72)} & \textbf{77.83}\tiny{(0.68)}\\
			3SADA+0Mix & \textbf{79.44}\tiny{(0.79)} & 80.68\tiny{(0.70)} & 70.77\tiny{(0.74)}& \textbf{78.42}\tiny{(0.68)} & 77.33\tiny{(0.62)}\\
			\hline
		\end{tabular}
	}
	\label{tab:CIFAR10_table}
\end{table}



\subsection{Model sensitivity perspective}
\label{sec:sensitivity_perspective}

\begin{figure}[t]
    \centering
    \includegraphics[width=\columnwidth]{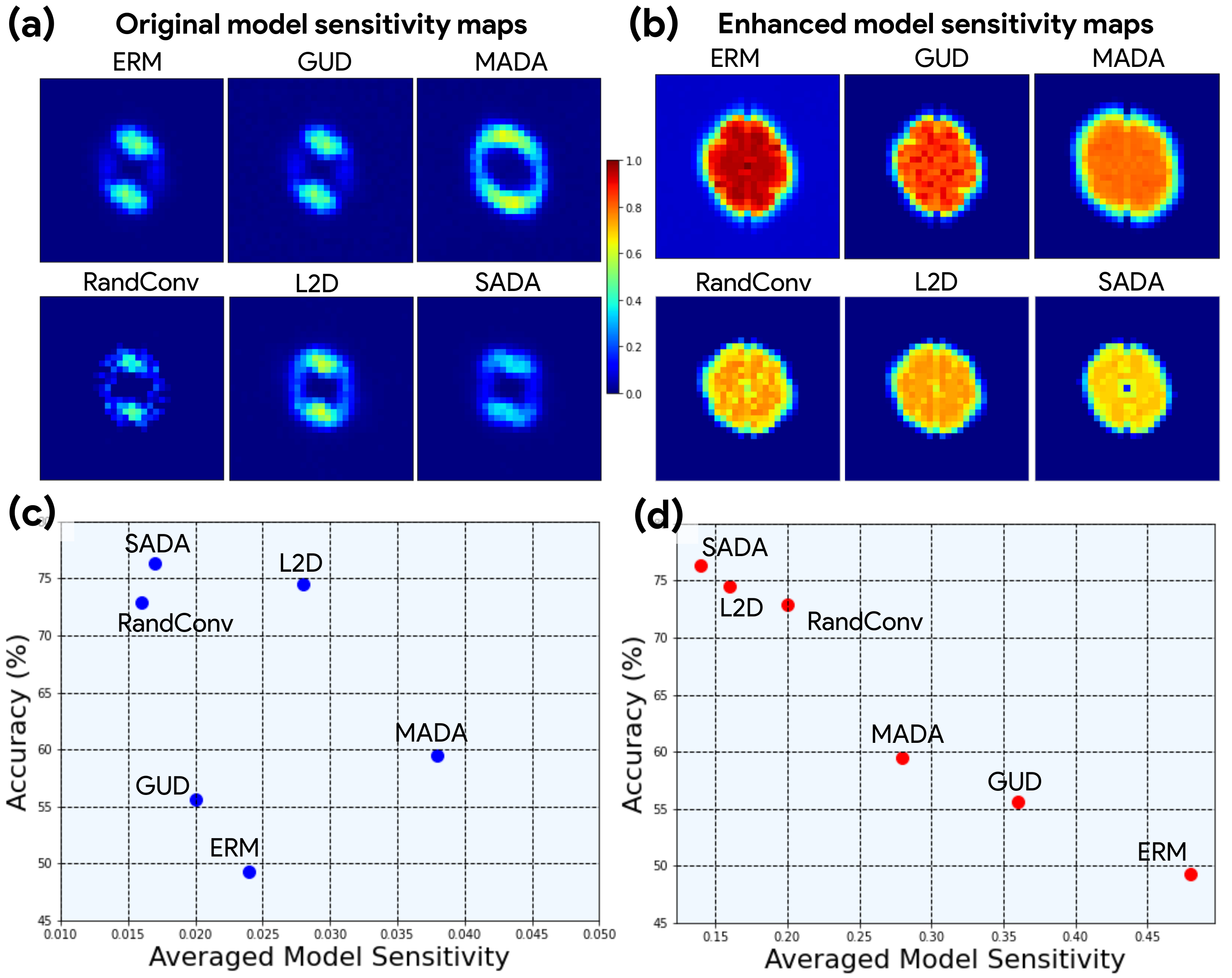}
    \caption{\textbf{(a)} and \textbf{(b)}: original and our proposed model sensitivity maps of different single-DG methods. \textbf{(c)} and \textbf{(d)}: model performance versus the model sensitivity maps.}
    \label{fig:sensitivity_analysis}
\end{figure}

To verify if the enhanced model sensitivity map can indicate the model’s generalizability, we further computed the sensitivity maps of ConvNet trained with different strategies on the \textbf{MNIST} dataset as shown by the examples in Fig.~\ref{fig:sensitivity_analysis}. 
%
\textit{First}, different from the original sensitivity maps (Eq.~\ref{eq:sensitivity_map_org}) in Fig.~\ref{fig:sensitivity_analysis} \textbf{(a)}, the enhanced sensitivity maps (Eq.~\ref{eq:sensitivity_map_enhanced}) in Fig.~\ref{fig:sensitivity_analysis} \textbf{(b)} show that source models are more vulnerable to the perturbations in the low frequency region. This result matches with the observations in the previous studies~\cite{yang2020fda,liu2021feddg}, that the models cannot generalize well due to the low-frequency amplitude difference between the source and target domains is large.
\textit{Second}, in Fig.~\ref{fig:sensitivity_analysis} \textbf{(b)}, comparing the model sensitivity map of ERM with the sensitivity maps of other single-DG approaches, our enhanced sensitivity computation clearly shows how the single-DG approaches can help improve model performance by suppressing the model sensitivity, especially in the low-frequency space.
%
To better visualize the observations, we present the scattering plot of the accuracy versus the averaged $\ell_1$-norm of model sensitivity map. As shown in Fig.~\ref{fig:sensitivity_analysis}\textbf{(c)}, the original sensitivity computation method fails to correlate the model performance and sensitivity. In contrast, Fig.~\ref{fig:sensitivity_analysis}\textbf{(d)} shows that the enhanced model sensitivity computation provides strong correlation between the model performance and sensitivity. The model prediction accuracy degrades significantly when the $\ell_1$-norm of model sensitivity maps increases.
These results demonstrate that the enhanced model sensitivity map in Eq.~\ref{eq:sensitivity_map_enhanced} could be used for \textit{visualizing and quantifying the effect of implicit regularization} on model generalizability.




\subsection{Data efficiency}
\label{sec:efficiency}

\begin{figure}[t]
    \centering
    \subfigure[DIGITS]{ \includegraphics[width=.48\columnwidth]{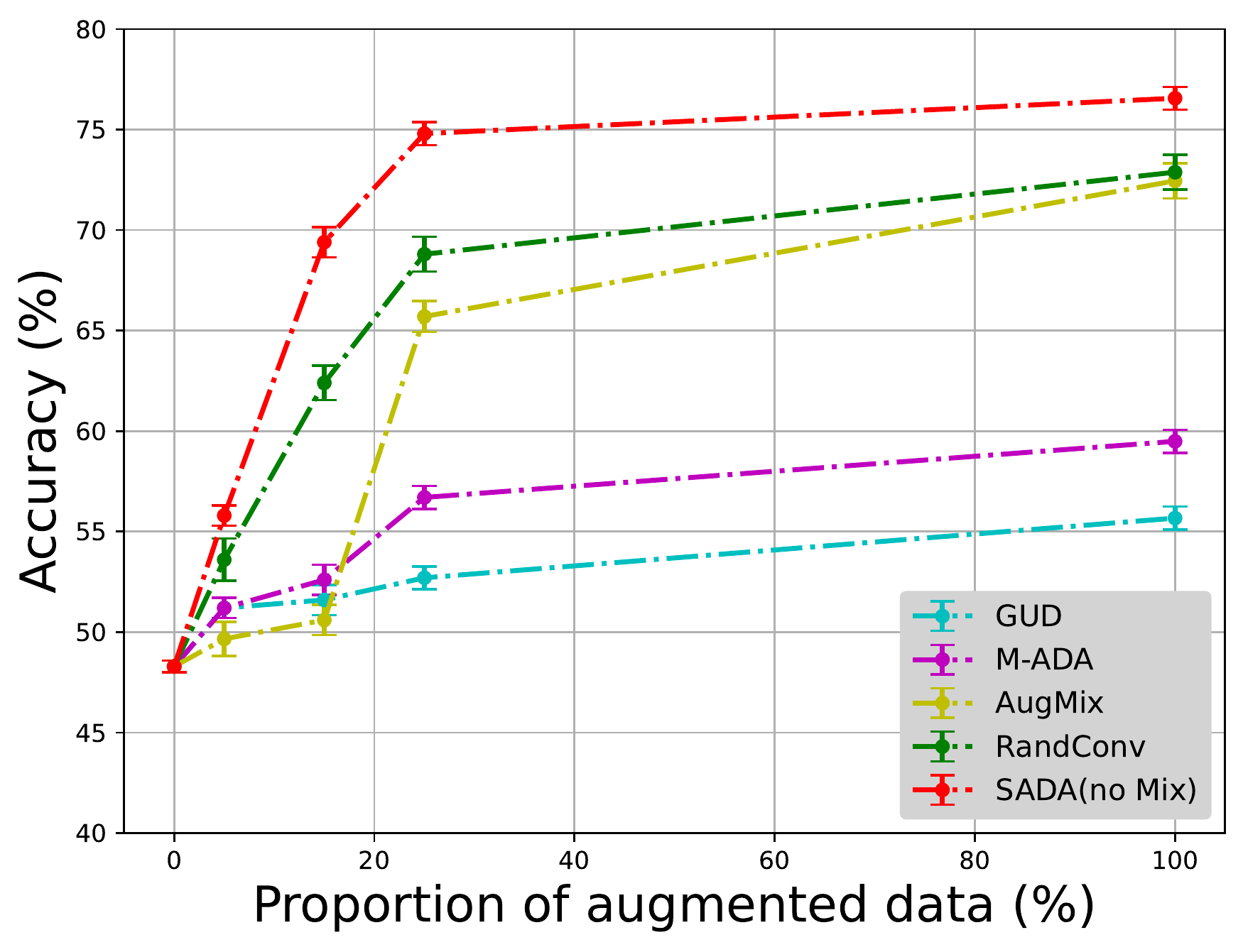}}
    \subfigure[PACS]{ \includegraphics[width=.48\columnwidth]{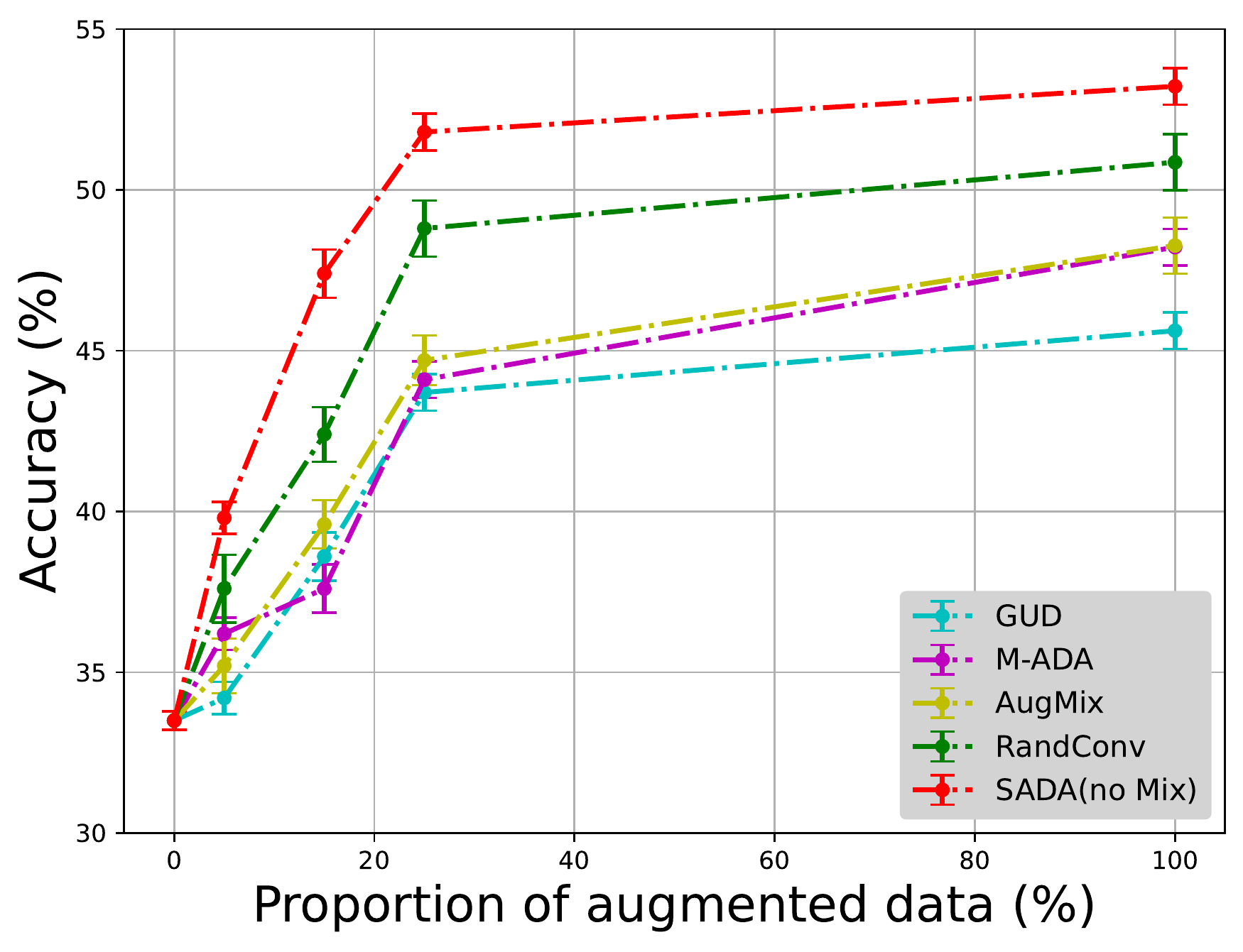}}
      
    \caption{Evaluation of data efficiency of SADA}
    \label{fig:data_efficiency}
\end{figure}

Due to the limited availability of data, data efficiency is an important indicator of data augmentation performance.
We evaluated the model accuracy by gradually decreasing the amount of augmented images used in the training process. Fig.~\ref{fig:data_efficiency} shows the analysis results of using \textbf{MNIST} and \textbf{P}hoto as source domains on \textbf{DIGITS} and \textbf{PACS}, respectively. Our method consistently outperforms other baselines when only a proportion of the augmented data are used for training. It is also very impressive to see that, with only $25\%$ augmented data for training, our method can generalize better than the baselines trained with the fully augmented dataset. 


\section{Further analysis and discussion}

\subsection{Ablation studies}

We conducted ablation studies on both \textbf{DIGITS} and \textbf{PACS} datasets to verify the effectiveness of each component in SADA. 
Table~\ref{tab:ablation_DIGITS} reports the performance with 2SADA+1Mix as example.
We first removed the two SADA augmented images (w/o SADA) and the performance degraded more than $20\%$ in all the target domains, which clearly shows the significance of SADA in the whole framework. 
Second, without the AugMix (w/o Mix), \textit{i.e.,} 2SADA+0Mix, the model performance decreased by $7\%$ in each target domain. That is because AugMix includes several random image style transfer operations, such as `solarization' and `autocontrast', which diversify the augmented images to complement our targeted spectral augmentation.
Third, we also observed the performance drop if the models are trained without the $\rm{JS}$ divergence, which helps learn invariant representations for improved generalizability.

\begin{table}[th]
	\centering
	\caption{Ablation of SADA(1Mix) on \textbf{DIGITS} dataset.} 
	\scalebox{0.7}{
		\begin{tabular}{|l||c|c|c|c||c|}
			\hline
			\textbf{Component} &  \textbf{USPS} & \textbf{MNIST-M} & \textbf{SVHN} & \textbf{SYNTH} & Avg\\
			\hline
			\hline

			2SADA+1Mix & 89.34  & 75.74 & 68.34 & 72.10 & 76.38 \\
			\hline
            w/o SADA & 69.62 & 53.57 & 47.16 & 49.02 & 60.27\\

            w/o Mix & 82.79 & 69.43 & 62.44 & 63.18 & 69.46\\
            
            w/o $\rm{JS}$ divergence & 81.68 & 70.35 & 64.32 & 65.11 & 70.37\\
			\hline
		\end{tabular}
	}
	\label{tab:ablation_DIGITS}
\end{table}

\subsection{Effectiveness of targeted augmentation}
\begin{figure}[bt]
    \centering
    \includegraphics[width=\columnwidth]{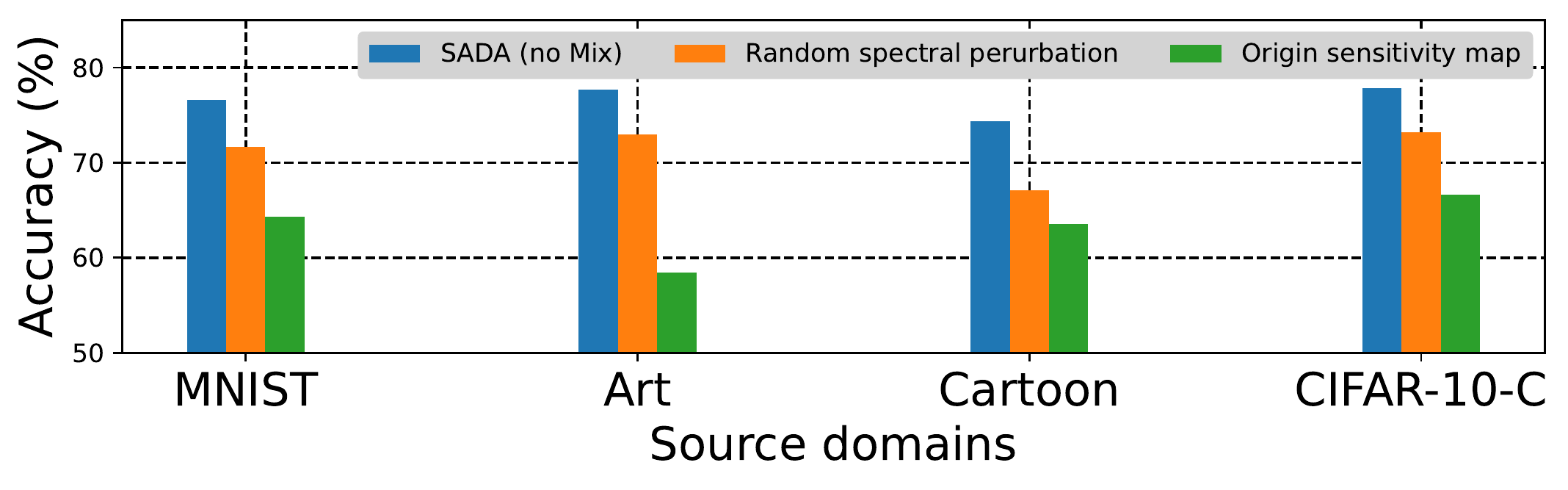}
    \caption{3SADA+0Mix performance comparison on model sensitivity map and spectral perturbation.}
    \label{fig:method_efficiency}
\end{figure}

This section examines the effectiveness of the proposed model sensitivity map and the targeted adversarial perturbation.
%
We evaluate the model performance with 3SADA+0Mix by $1)$ using the original model sensitivity map; $2)$ replacing the adversarial spectral augmentation with the random spectral perturbation($\epsilon=0.2$) following~\cite{sun2021certified}.
The results in Fig.~\ref{fig:method_efficiency} show that 3SADA+0Mix generalizes worse to unseen domains if our proposed components are replaced by the two alternative approaches on different source domains.
In addition, we also evaluate the time consumption regarding the sensitivity map generation and the spectral adversarial augmentation in the appendix.

\subsection{Hyperparameter sensitivity analysis}
\label{sec:parameter}
\begin{figure}[tb]
    \centering
    \includegraphics[width=\columnwidth]{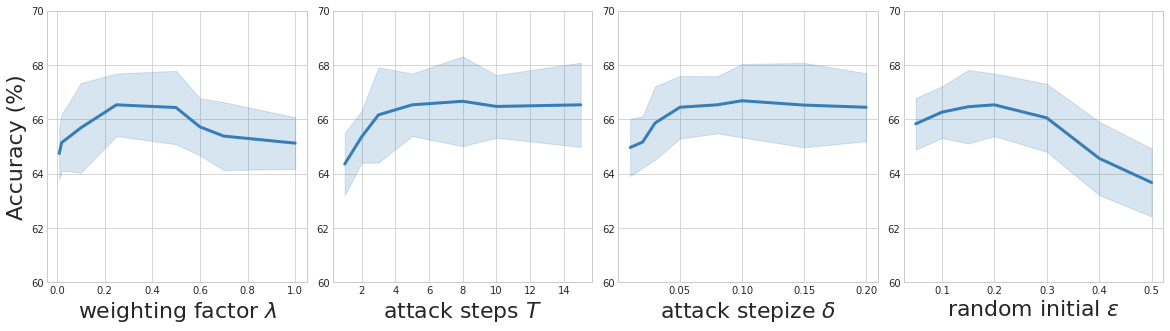}    \caption{Sensitivity analysis of different hyperparameters.}
    \label{fig:params_sensitivity}
\end{figure}

To validate the significance of weighting factor $\lambda$, perturbation steps $T$ and step size $\delta$, and random initialization $\epsilon$, we conduct sensitivity analysis of 2SADA+1Mix on \textbf{PACS} dataset as presented in Fig.~\ref{fig:params_sensitivity}. In the experiments, we initially set $\lambda=0.25$, $T=5$, $\delta=0.08$ and $\epsilon=0.20$. When analyzing the sensitivity to one parameter, the other parameters are fixed. When $\lambda$ is within $[0.1, 0.6]$, our method consistently outperforms other baselines (Fig.~\ref{fig:params_sensitivity}\textbf{a)}). That is due to the balance between the $\rm{JS}$ loss and the ERM loss. When the perturbation gets stronger, Fig.~\ref{fig:params_sensitivity}\textbf{b)} and \textbf{c)} show that the performance increases initially, and then stays stable. It is because the early-stop acceleration is adopted to control the perturbation strength. Fig.~\ref{fig:params_sensitivity}\textbf{d)} shows that the model performance is stable if the perturbation strength $\epsilon < 0.30$, which decreases if the randomization gets too strong.

\section{Conclusion and Discussion}

In this paper, an enhanced model sensitivity map is proposed to empirically ascertain \emph{when the neural networks may fail in domain generalization} from a new perspective of spectral sensitivity. Our analysis shows that models with high sensitivity may not generalize well. Based on our analysis, we develop a novel framework of \emph{Spectral Adversarial Data Augmentation} (SADA) to tackle single-DG by generating adversarially augmented images targeted at the highly sensitive frequencies.
By successfully suppressing the model sensitivity in the frequency space, the experimental results on three public benchmarking datasets demonstrate that SADA can efficiently train a high-performance model resilient to various unseen domain shifts.

A limitation of our work is that we mainly target at the sensitive regions of source domain models. While this is very efficient for suppressing the model sensitivity, it can be less effective when the domain shift spreads over the frequency space. 
In conclusion, the effectiveness of SADA and the revealed intrinsic correlation between model generalizability and sensitivity may have a profound influence on the research of domain generalization and adaptation.

\section*{Acknowledgement}
This research was partially supported by the National Science Foundation (NSF) under the CAREER award OAC 2046708 and the Rensselaer-IBM AI Research Collaboration (http://airc.rpi.edu), part of the IBM AI Horizons Network (http://ibm.biz/AIHorizons).

\bibliography{aaai23}

\end{document}